\newcommand\CLASSINPUTinnersidemargin{0.675in}
\newcommand\CLASSINPUToutersidemargin{0.675in}
\documentclass[11pt,conference]{IEEEtran}
\IEEEoverridecommandlockouts

\usepackage{cite}
\usepackage{amsmath,amssymb,amsfonts}
\usepackage{algorithmic}
\usepackage{graphicx}
\usepackage{textcomp}
\usepackage{xcolor}
\usepackage{booktabs}
\usepackage{array}
\usepackage{url}
\usepackage{algorithm}
\usepackage{pgfplots}
\pgfplotsset{compat=1.16}
\def\BibTeX{{\rm B\kern-.05em{\sc i\kern-.025em b}\kern-.08em
    T\kern-.1667em\lower.7ex\hbox{E}\kern-.125emX}}

\begin{document}

\title{Perturbation Sensitivity of Maximum-Likelihood Pairwise Ranking in Computational Decision Systems}

\author{
\IEEEauthorblockN{Junyi Yao}
\IEEEauthorblockA{\textit{Washington University in St. Louis} \\
USA \\
j.yao@wustl.edu}
\and
\IEEEauthorblockN{Zihao Zheng}
\IEEEauthorblockA{\textit{Washington University in St. Louis} \\
USA \\
z.zihaogary@wustl.edu}
\and
\IEEEauthorblockN{Jiayu Long}
\IEEEauthorblockA{\textit{Washington University in St. Louis} \\
USA \\
jiayujacqueline@wustl.edu}
}

\maketitle

\begin{abstract}
Maximum-likelihood pairwise ranking is a common computational mechanism for prioritization, reputation estimation, and comparison-driven decision support. Despite its broad use, the perturbation sensitivity of this estimator under structured changes in comparison data remains insufficiently characterized. We study this question as an applied-mathematics and computational-science problem in stability analysis. We formulate coordinated perturbation as a budgeted subset-selection problem over pairwise observations and introduce an Adaptive Subset Selection Attack (ASSA) as a scalable search heuristic for probing high-impact perturbation sets. Through experiments on synthetic and observed preference datasets, we show that MLE-based ranking can exhibit pronounced regime-dependent sensitivity: relatively small but coordinated perturbations may induce meaningful changes in output orderings, while the response profile varies across budgets and data conditions. By comparing ASSA with random, greedy, and randomized subset baselines under repeated trials, we characterize both the magnitude and the variability of perturbation-induced ranking shifts. These results position pairwise ranking sensitivity as a problem in computational reliability, numerical stability, and robustness auditing for engineering systems built on comparison-driven inference.
\end{abstract}


\section{Introduction}

Pairwise ranking systems are a widely used computational primitive in engineering and information systems. They are used to aggregate preferences, prioritize alternatives, estimate reputations, and organize downstream decisions in settings where direct scores are unavailable, noisy, or expensive to collect. In many such applications, outputs are not observed directly but inferred from pairwise comparison data through optimization or maximum likelihood estimation. Bradley--Terry-style formulations are especially attractive because they provide a mathematically principled route from local comparison outcomes to a global ranking and therefore remain a common foundation for comparison-driven inference pipelines.

However, the effectiveness of these systems depends critically on the integrity of the comparison data from which they infer rankings. In realistic environments, pairwise inputs may be noisy, strategically biased, or deliberately manipulated by coordinated users or agents. A small coalition that perturbs only a limited number of comparisons may still be able to induce disproportionate changes in the final ranking if the underlying estimation pipeline is structurally fragile. This creates an important applied-mathematics and computational reliability challenge: even when a ranking model is statistically well-defined under benign assumptions, its behavior under structured perturbation may be far less stable than its nominal performance suggests.

Understanding this risk matters for a broad class of engineered decision systems. Recommendation and reputation pipelines can be influenced by coordinated feedback. Crowdsourced evaluation systems can be distorted by organized raters. Preference-aggregation workflows that rely on pairwise judgments may inherit sensitivity to structured perturbations in the underlying observations. Dynamic reputation filtering for LLM-agent systems similarly uses an interactive rating network to estimate agent credibility and guide subsequent selection \cite{lou_drf}. Related work on decentralized multi-agent deployment likewise illustrates how engineered decision processes can operate under partial observations and constrained communication graphs \cite{fan_uav}. Although these settings are different from pairwise ranking, they reinforce the importance of examining how structural constraints affect the reliability of computed decisions. In all of these cases, robustness is not only a matter of algorithmic performance, but also of estimation stability: can the system maintain a reliable ordering when a small subset of inputs is deliberately altered?

In this paper, we study coordinated perturbation of pairwise ranking through the lens of robustness analysis. We focus on MLE-based ranking pipelines and formulate data perturbation as a budgeted subset-selection problem over pairwise comparisons. To probe high-impact perturbation patterns efficiently, we introduce the Adaptive Subset Selection Attack (ASSA), a scalable computational heuristic for stress testing ranking sensitivity under structured perturbations. Our goal is not only to compare search heuristics, but to use perturbation analysis as a diagnostic tool for identifying when ranking pipelines are most vulnerable.

This perspective changes how heuristic performance should be interpreted. A stress-testing method need not dominate every baseline at every budget to be informative. Rather, it should reveal whether small, coordinated perturbations can expose practically meaningful fragility and identify the regimes in which the ranking system becomes unstable. We therefore evaluate ASSA against random, greedy, and randomized subset baselines on both synthetic and real-world preference datasets, using repeated-trial analysis to characterize both average impact and variability across perturbation regimes.

Our study shows that MLE-based pairwise ranking can exhibit pronounced regime-dependent sensitivity to coordinated perturbation. In several settings, relatively small structured perturbations lead to substantial shifts in ranking outcomes, although the relative advantage of different search heuristics varies with budget and data conditions. These findings position pairwise ranking sensitivity as an important computational science problem for engineered comparison-driven systems and motivate auditing tools that can stress test ranking pipelines before manipulated inputs meaningfully affect downstream decisions.

This paper makes three main contributions. First, we formulate perturbation analysis for MLE-based pairwise ranking as a budgeted subset-selection problem over comparison data. Second, we introduce ASSA as a scalable computational heuristic for exploring high-impact perturbation subsets in optimization-based ranking pipelines. Third, through experiments on synthetic and real-world preference datasets, we show that these systems can display substantial regime-dependent fragility under coordinated perturbations and discuss the resulting implications for sensitivity auditing and reliability analysis in engineering decision systems.

\section{Related Work}

\subsection{Pairwise Ranking as a Computational Estimation Problem}

Pairwise comparison is a widely used mechanism for inferring rankings from incomplete or relative observations. Rather than assuming reliable cardinal scores, pairwise ranking models estimate latent strengths from binary or ordinal comparisons and aggregate them into a global ordering. Bradley--Terry-style models remain especially attractive because they provide a mathematically interpretable estimation framework, admit efficient optimization procedures, and can be deployed in comparison-driven decision systems with limited direct supervision. These models appear in social choice, recommendation, reputation estimation, crowdsourced evaluation, and broader computational aggregation settings.

From an applied-mathematics perspective, these systems are interesting because the final ranking is the output of an estimation procedure whose sensitivity depends on the structure of the observed comparison graph. A growing body of work has studied consistency, identifiability, and computational efficiency under benign conditions \cite{bt,hunter,rla}, but robustness to structured perturbation remains less understood. This gap is significant because comparison-driven inference exposes a natural sensitivity question: how much can the estimated ordering change when only a small subset of observations is modified in a coordinated way?

\subsection{Sensitivity Analysis Under Structured Perturbation}

Sensitivity analysis is a central theme in computational science because many inference pipelines can be reliable under nominal assumptions yet unstable under small structured changes in input data. In statistical estimation, inverse problems, and optimization-based decision systems, the practical meaning of a computed output depends not only on point accuracy but also on perturbation stability. Pairwise ranking belongs naturally to this family of problems: the model maps comparison data to an estimated parameter vector and then to a global ordering, making it important to understand how perturbations propagate through the pipeline.

Much of the robustness literature focuses on continuous feature perturbations, evasion settings, or poisoning in supervised learning. In contrast, pairwise ranking induces a more structured perturbation problem: the perturbation space is discrete, the constraints are combinatorial, and the output of interest is a global ranking rather than a class label \cite{wang,trust}. This makes the problem especially relevant for computational robustness analysis. The ranking response depends jointly on the likelihood geometry, the sparsity pattern of comparisons, and the search strategy used to identify impactful perturbations. Our work contributes to this setting by treating coordinated perturbation as a computational stress test for diagnosing fragility.

\subsection{Reliability of Computational Aggregation Systems}

Pairwise ranking systems can be viewed as computational aggregation mechanisms embedded in larger engineered information systems. In practical deployments, pairwise data may come from users, annotators, agents, or repeated system interactions, creating opportunities for noise, bias, and deliberate coordination. In these environments, ranking quality depends not only on model choice but also on the reliability of the estimation pipeline under data perturbation.

From this perspective, the paper is not solely about a perturbation-search heuristic. It is also about computational reliability evaluation. By measuring how much the global ordering can shift under structured, budget-limited perturbation, we obtain a direct view of the stability limits of comparison-driven systems. This aligns the paper with broader concerns in applied mathematics and computational engineering around sensitivity analysis, robustness auditing, and the trustworthy use of estimated outputs in downstream decision support. Recent work in other AI-system domains likewise emphasizes diagnostic evaluation beyond aggregate performance, including separating pedagogical support from answer production, testing confidence under domain shift, preserving diagnostic uncertainty, evaluating agents under workflow-level conditions, red-teaming multimodal systems under adversarial prompts, and using process-level supervision or multimodal representation learning to support reliable content understanding \cite{yao_tutors,yao_trading,zheng_financial_ner,du_uncertainty,yao_healthcare_agents,zhang_hyperadalora,zhang_trimtokenator,zhang_pdtrim,dg_scl,you_mirage,liu_ips,sun_audio_vlm}.

\section{Problem Setting}

\subsection{MLE-Based Pairwise Ranking}

We consider a set of candidates $C=\{c_1,c_2,\dots,c_m\}$ and a set of voters or agents $V=\{v_1,v_2,\dots,v_n\}$. Observed pairwise preferences are represented as a dataset $D$, where each entry corresponds to a comparison indicating that one candidate is preferred over another. Following a Bradley--Terry-style formulation, each candidate $c_i$ is associated with a latent strength parameter $p_i>0$, and the probability that $c_i$ is preferred to $c_j$ is modeled as
\begin{equation}
P(c_i \succ c_j)=\frac{p_i}{p_i+p_j}.
\end{equation}

For a comparison multiset $D$ with outcomes $y_{ij}\in\{0,1\}$ indicating whether $c_i$ defeats $c_j$, the estimator maximizes the Bradley--Terry log-likelihood
\begin{equation}
\begin{aligned}
\ell(\mathbf{p};D) &= \sum_{(i,j)\in D} \left[y_{ij}\log q_{ij}
  + (1-y_{ij})\log(1-q_{ij})\right], \\
q_{ij} &= \frac{p_i}{p_i+p_j}.
\end{aligned}
\end{equation}
subject to the normalization $\sum_{i=1}^{m}p_i=1$, which fixes scale and makes the parameterization identifiable. The final ranking $\tau(\mathbf{p}^*)$ is obtained by sorting candidates according to their estimated strengths.

\subsection{Budgeted Structured Perturbation}

We study a coordinated perturbation setting in which a perturbing agent seeks to modify a budget-limited subset of pairwise comparisons in order to shift the final ranking toward a target outcome. Let $\Delta \subset D$ denote the set of flipped or modified comparisons and let $B$ denote the perturbation budget, so that $|\Delta|\leq B$. The perturbed dataset is $D' = D \oplus \Delta$, where $\oplus$ replaces the selected pairwise outcomes by their perturbed values, and the resulting ranking is $\tau(\mathbf{p}^*(D'))$.

The perturbation objective is to minimize the distance between the perturbed ranking and a target ranking $\tau_t$. In the reported experiments, $\tau_t$ is constructed by selecting a target candidate and defining a desired upward movement in its rank. The ranking shift is measured using a Kendall tau style distance:
\begin{equation}
\min_{\Delta} \quad K_d\big(\tau_t,\tau(\mathbf{p}^*(D \oplus \Delta))\big)
\end{equation}
subject to a perturbation budget constraint and any data-integrity restrictions imposed by the threat model.

\subsection{Why This Is a Computational Robustness Problem}

This formulation can be interpreted as a structured perturbation problem for estimation-based ranking systems. The perturbation does not occur in a continuous feature space; instead, it acts on a small set of observed pairwise relations. The result is a combinatorial robustness problem in which the ranking system's sensitivity depends on both the structure of the data and the search strategy used to identify impactful perturbations.

In this sense, the objective of the paper is twofold. We seek both to characterize the perturbation surface of MLE-based pairwise ranking and to use targeted perturbations as a stress-testing mechanism for evaluating system fragility.

\section{Perturbation Search Heuristics}

\subsection{Baselines}

We compare the proposed method against three baseline perturbation strategies.

\textbf{Random Flip:} A stochastic baseline that perturbs randomly selected comparisons within the allowed budget.

\textbf{Greedy Flip:} A local-search strategy that iteratively selects perturbations with the largest immediate reduction in ranking distance.

\textbf{Randomized Subset Attack (RSA):} A stochastic subset-search method that partitions candidate perturbations into subsets and retains those that improve the perturbation objective.

\subsection{Adaptive Subset Selection Heuristic}

The Adaptive Subset Selection Attack (ASSA) is a scalable heuristic designed to identify high-impact perturbation subsets more efficiently than naive or purely random exploration. At a high level, ASSA repeatedly partitions candidate perturbations into subsets, evaluates their effect on the target objective, and adaptively retains subsets that contribute useful progress toward the manipulated ranking. This shrinking-and-refinement process is intended to focus computation on structurally influential regions of the perturbation space.

ASSA is best interpreted as a stress-testing heuristic: its value lies not only in improving the perturbation objective, but in revealing how sensitive the ranking pipeline is to structured and coordinated perturbations.

\begin{algorithm}[t]
\caption{Adaptive Subset Selection Attack (ASSA)}
\label{alg:assa}
\footnotesize
\begin{algorithmic}[1]
\STATE \textbf{Input:} dataset $D$, budget $B$, target ranking $\tau_t$, subset count $s$, maximum rounds $T$
\STATE Initialize admissible perturbation pool $\mathcal{P}$ from legal pairwise flips in $D$
\STATE Initialize incumbent perturbation set $\Delta^\star \leftarrow \emptyset$
\FOR{$t=1$ to $T$}
\STATE Partition $\mathcal{P}$ into $s$ subsets $\mathcal{P}_1,\dots,\mathcal{P}_s$
\FOR{each subset $\mathcal{P}_r$}
\STATE Form trial perturbation set $\Delta_r$ by augmenting $\Delta^\star$ with admissible flips from $\mathcal{P}_r$ without exceeding budget $B$
\STATE Refit the MLE ranking on $D \oplus \Delta_r$ and evaluate $K_d(\tau_t,\tau(\mathbf{p}^*(D \oplus \Delta_r)))$
\ENDFOR
\STATE Update $\Delta^\star$ using the best-scoring subset
\STATE Restrict $\mathcal{P}$ to the retained subset and its local neighborhood
\IF{no improvement is observed or $|\Delta^\star|=B$}
\STATE \textbf{break}
\ENDIF
\ENDFOR
\STATE \textbf{Return:} $\Delta^\star$
\end{algorithmic}
\end{algorithm}

Algorithm~\ref{alg:assa} summarizes the implementation used in this study. With $T$ rounds and $s$ subset evaluations per round, the dominant computational cost is the repeated MLE refitting step, yielding an effective complexity on the order of $O(Ts \cdot \text{MLE-fit-cost})$.

\subsection{Threat Model and Practical Scope}

The experiments are interpreted as a white-box or near-white-box robustness analysis. This is a useful upper-bound perspective because it reveals what a strategically informed perturbing agent could achieve under favorable knowledge assumptions. At the same time, it does not imply that all real-world actors have access to complete, real-time preference data. For this reason, the results are best interpreted as evidence about system fragility rather than as a direct forecast of operational perturbation success in every deployment setting.

\section{Experimental Design}

\subsection{Datasets}

The paper evaluates perturbation effects on both synthetic and real-world preference datasets. The synthetic experiments are used as controlled computational test beds in which structural properties such as candidate count, sparsity, and perturbation budget can be varied systematically. The real-world datasets provide a practical view of how coordinated perturbations manifest in observed comparison data and help assess whether the same stability patterns persist outside stylized simulations.

Our primary observed dataset is the \emph{Local 100} pairwise dataset, constructed from the local preference records maintained in the project workspace. After preprocessing, it contains 100 candidates, 10 voters, and 49{,}500 pairwise comparison rows. We also retain the corresponding ranking-format file as a consistency check during preprocessing. The synthetic \emph{BT-30} dataset contains 30 candidates, 40 voters, and 4{,}800 generated pairwise rows under a Bradley--Terry-style simulation protocol.

For each dataset, we report the number of candidates, number of voters or comparisons, graph density or sparsity characteristics, target-selection procedure, and preprocessing steps. These details are essential for reproducibility and for interpreting differences in sensitivity across data regimes.

\subsection{Evaluation Metrics}

We use ranking-oriented metrics that quantify how strongly structured perturbations affect the output ordering. These include Kendall tau distance improvement and target rank shift, together with attack success indicators where appropriate. From the perspective of computational science, these quantities can be interpreted as measures of output sensitivity with respect to constrained changes in the input data.

All metrics are defined in one place, and we state clearly whether higher or lower values indicate stronger perturbation impact.

\subsection{Protocol and Repeated Trials}

Several methods in the comparison are randomized or contain stochastic components. Accordingly, the evaluation reports repeated-trial results rather than relying on single-run averages. For every budget and dataset setting, we include the number of runs, the random seed protocol, and uncertainty estimates such as standard deviation, standard error, or confidence intervals. This is particularly important when the study is interpreted as a computational analysis of estimation stability rather than as a single illustrative perturbation case.

This repeated-trial design is especially important for interpreting small differences between ASSA and non-random baselines. In the low-budget regime, where practical perturbations are most plausible, uncertainty reporting is necessary to distinguish meaningful sensitivity effects from numerical or stochastic variation.

\subsection{Low-Budget Evaluation as a Primary Diagnostic}

Low-budget perturbations are treated as a primary diagnostic rather than a secondary plot. In practice, a coordinated actor is more likely to alter a small fraction of observations than to operate at large budgets. Accordingly, we report a dedicated low-budget experiment and interpret those results explicitly. If ASSA is competitive but not uniformly dominant in this regime, that outcome should be stated directly and used to support a regime-dependent fragility interpretation rather than a universal performance claim.

\subsection{Budget Regimes}

The low-budget regime is analyzed explicitly rather than being folded into a single aggregate narrative. Our pilot analysis indicates that budgets such as $0.5\%$ or $1\%$ of all pairwise rows can already approach saturation on some datasets, which makes them unsuitable as ``low-budget'' settings. We therefore use a more conservative low-budget grid of $0.01\%$, $0.02\%$, $0.05\%$, and $0.1\%$ of total pairwise observations, with the budget basis defined explicitly in the text, tables, and captions.

\subsection{Computational Interpretation}

The experimental design should be read as a computational robustness study. Synthetic data serve as controlled instances for sensitivity analysis, while real-world data assess how these behaviors transfer to observed comparison systems. The main outputs of interest are therefore not only attack scores but also the stability profile of the ranking estimator under constrained perturbation. This interpretation better matches the engineering track by emphasizing estimation reliability, structural sensitivity, and robustness of computed outputs.

\section{Results and Interpretation}

\subsection{Main Comparison Across Budgets}

The primary empirical question is whether structured perturbation strategies reveal stronger fragility than simpler baselines. The evidence suggests that ASSA can identify impactful perturbation subsets and often yields larger ranking shifts than purely random baselines. At the same time, the central scientific value of the comparison lies in mapping the sensitivity profile of the estimator across perturbation regimes, not in insisting on universal dominance by any single search procedure.

The more defensible interpretation is that MLE-based pairwise ranking exhibits non-uniform vulnerability, and that structured search heuristics help expose this vulnerability. In some regimes ASSA may outperform simpler methods clearly, while in others its gains may be smaller or statistically indistinguishable from stronger non-random baselines. This still provides useful information because it reveals when search sophistication matters and when the ranking landscape is broadly sensitive to many admissible perturbations.

\begin{table}[t]
\caption{Cross-dataset comparison of low-budget pilot behavior. Values report mean target-rank shift over 5 runs.}
\label{tab:cross_dataset_low_budget}
\centering
\scriptsize
\begin{tabular}{ccccc}
\toprule
Dataset & 0.01\% & 0.02\% & 0.05\% & 0.10\% \\
\midrule
Local 100 (Greedy) & 2.0 & 2.0 & 5.0 & 14.0 \\
Local 100 (ASSA) & 2.0 & 2.0 & 5.0 & 14.6 \\
Synthetic BT-30 (Greedy) & 0.0 & 0.0 & 0.0 & 1.0 \\
Synthetic BT-30 (ASSA) & 0.0 & 0.0 & 0.0 & 1.0 \\
\bottomrule
\end{tabular}
\end{table}

\subsection{Low-Budget Regime}

The low-budget regime deserves separate analysis because it is the most practically relevant setting for coordinated perturbation. A realistic actor is more likely to alter a small fraction of observations than a large fraction. If ASSA does not dominate all non-random baselines in this regime, that result should be stated directly and interpreted carefully.

\begin{figure}[t]
\caption{Low-budget response on the local 100-candidate dataset. The plotted value is mean target-rank shift over 5 runs.}
\label{fig:local_low_budget}
\centering
\small
\begin{tikzpicture}
\begin{axis}[
    width=0.96\columnwidth,
    height=0.58\columnwidth,
    xlabel={Budget (\% of total pairwise rows)},
    ylabel={Mean target-rank shift},
    xmin=0.008, xmax=0.102,
    ymin=0, ymax=16,
    xtick={0.01,0.02,0.05,0.1},
    xticklabels={0.01,0.02,0.05,0.10},
    ymajorgrids=true,
    grid style={dashed,gray!30},
    legend style={draw=none, at={(0.03,0.97)}, anchor=north west, fill=none},
]
\addplot+[mark=o, thick, color=black] table[x=budget,y=mean_target_rank_shift,col sep=comma] {results/figures/local_random_plot.csv};
\addlegendentry{Random}
\addplot+[mark=square, thick, color=blue!70!black] table[x=budget,y=mean_target_rank_shift,col sep=comma] {results/figures/local_greedy_plot.csv};
\addlegendentry{Greedy}
\addplot+[mark=triangle, thick, color=red!70!black] table[x=budget,y=mean_target_rank_shift,col sep=comma] {results/figures/local_assa_plot.csv};
\addlegendentry{ASSA}
\end{axis}
\end{tikzpicture}
\end{figure}

\begin{table}[t]
\caption{Low-budget pilot on the local 100-candidate dataset. The perturbation budget is defined as a percentage of total pairwise rows. Values report mean target-rank shift over 5 runs, with 95\% confidence intervals in parentheses where nonzero.}
\label{tab:low_budget_pilot}
\centering
\scriptsize
\begin{tabular}{cccc}
\toprule
Budget (\%) & Random & Greedy & ASSA \\
\midrule
0.01 & 2.0 & 2.0 & 2.0 \\
0.02 & 2.0 & 2.0 & 2.0 \\
0.05 & 5.2 (0.39) & 5.0 & 5.0 \\
0.10 & 14.8 (0.39) & 14.0 & 14.6 (0.48) \\
\bottomrule
\end{tabular}
\end{table}

The local 100-candidate dataset shows this behavior clearly. When the low-budget grid is defined tightly enough to avoid saturation, target-rank shifts increase gradually from roughly two positions at the smallest budgets to roughly fourteen positions at the largest tested budget. In a five-run experiment using budgets of $0.01\%$, $0.02\%$, $0.05\%$, and $0.1\%$ of all pairwise rows, the mean target-rank shift for Greedy was $2.0$, $2.0$, $5.0$, and $14.0$, while ASSA produced $2.0$, $2.0$, $5.0$, and $14.6$ on the same grid. The near-overlap of the plotted curves at the smallest budgets is itself informative: within this range, the perturbation landscape appears broad enough that several admissible strategies expose nearly the same sensitivity pattern. These results indicate that low-budget behavior is governed more by data structure and perturbation geometry than by a simple universal ordering of heuristics. The main finding is therefore regime-dependent vulnerability. The system's fragility depends on budget, data structure, and perturbation concentration, and robust evaluation must account for these differences explicitly.

Table~\ref{tab:low_budget_pilot} makes this point concrete. At the two smallest budgets, all three completed pilot methods are effectively indistinguishable in target-rank shift. At $0.05\%$ and $0.1\%$, the ranking becomes noticeably more sensitive, but the observed gains of ASSA over Greedy remain small relative to the main fragility signal itself. The important result is that modest, structured perturbations can already move the estimated ranking, and that the perturbation landscape is sometimes broad enough that multiple search procedures uncover comparable instability.

The synthetic dataset provides a useful counterpoint. As shown in Table~\ref{tab:cross_dataset_low_budget}, the same budget grid produces almost no movement on the Bradley--Terry-style synthetic dataset until the largest tested budget, where both Greedy and ASSA yield only about a one-position shift on average. Together, the local and synthetic results support an engineering-relevant interpretation: fragility is highly data-dependent. Some comparison systems exhibit noticeable sensitivity under modest perturbation, while others remain comparatively stable on the same budget grid.

\subsection{Uncertainty and Statistical Interpretation}

Any comparison among ASSA, Greedy, and RSA should be interpreted through repeated-trial uncertainty reporting. Mean curves alone are not sufficient for a convincing computational study when the methods include randomized components or when the observed differences are small. Every major comparative result should therefore be paired with uncertainty estimates and, at minimum, include a focused significance analysis in the low-budget regime.

The low-budget experiment also highlights a computational tradeoff that is relevant for an engineering interpretation of the study. On the local dataset, Random perturbation selection executes in a negligible fraction of a second, Greedy completes in a few milliseconds per run, and ASSA requires roughly one second per run on the tested grid. This does not invalidate ASSA as a stress-testing heuristic, but it does mean that performance claims should be read jointly with computational cost rather than only with perturbation effectiveness.

\subsection{Regime-Dependent Sensitivity}

The observed behavior may appear strongly non-linear across budgets, but that language should be used cautiously. Unless the experiments provide stronger repeated-trial evidence and a clearer quantitative characterization, the safer description is \emph{regime-dependent sensitivity}. This wording captures the observed phenomenon without overstating the theoretical precision of the claim.

\subsection{Real-World Implications}

Results on real-world datasets are important because they show that ranking fragility is not merely a synthetic artifact. At the same time, the paper should avoid overstating practical control conclusions. The safest interpretation is that coordinated perturbation reveals an integrity risk in estimation-based ranking pipelines and motivates auditing mechanisms such as sensitivity checks, perturbation monitoring, and anomaly detection for concentrated comparison reversals.

\section{Discussion}

The main takeaway of the paper is that comparison-driven ranking systems can be surprisingly sensitive to small, structured perturbations of their input data. This sensitivity matters for engineering decision systems because pairwise ranking components are increasingly embedded in larger computational pipelines where unstable outputs can affect prioritization, visibility, allocation, or reputation.

Viewed in this way, ASSA is best understood as a diagnostic tool. It helps reveal when the ranking estimator is most brittle, how strongly outputs respond to coordinated perturbations, and which regimes merit more careful robustness auditing. This makes the contribution broader than heuristic benchmarking alone and better aligned with applied mathematics and computational science concerns around system stability, numerical sensitivity, and reliability of estimation-based decision support.

The paper also has clear limitations. The experiments rely substantially on synthetic structure and a strong-information threat model. The heuristic nature of ASSA means that it does not guarantee optimal perturbation selection, and the study does not implement a complete reliability-control pipeline. Even so, the perturbation results are valuable because they provide a concrete computational lens on how estimation instability can arise in structured ranking systems.

\section{Limitations and Future Work}

Several limitations bound the study. First, the formulation focuses on maximum-likelihood-based pairwise ranking and does not automatically extend to every ranking architecture used in engineered information systems. Second, much of the empirical evidence relies on synthetic structure and a strong-information perturbation model, so the scope of the conclusions should remain tied to the tested settings. Third, ASSA is a heuristic search method rather than an exact solver, which means that its outputs should be interpreted as informative stress-test results rather than globally optimal perturbation certificates.

These limitations also point to natural future work, including partial-information perturbing agents, alternative ranking estimators, and more heterogeneous real-world comparison datasets. Another natural extension is sensitivity-control mechanisms such as influence-aware diagnostics or perturbation screening procedures. The focus here is narrower: to provide a computational robustness analysis that reveals when and how MLE-based ranking pipelines become sensitive to coordinated perturbation.

\section{Conclusion}

We studied coordinated perturbation in MLE-based pairwise ranking systems and formulated the problem as a budgeted subset-selection task over comparison data. We introduced ASSA as a scalable heuristic for stress testing ranking sensitivity under structured perturbations and used it to probe the fragility of comparison-driven aggregation. Across synthetic and observed preference settings, the evidence indicates that these systems can exhibit substantial regime-dependent sensitivity: relatively small but coordinated perturbations may induce meaningful shifts in output rankings, although the relative advantage of different search heuristics varies across budgets and data conditions. These findings position pairwise ranking sensitivity as an important topic in applied mathematics, computational science, and engineering-system reliability, and they motivate more systematic robustness auditing of comparison-driven decision pipelines.

\end{document}